\documentclass[sigconf]{acmart}

\AtBeginDocument{%
  \providecommand\BibTeX{{%
    \normalfont B\kern-0.5em{\scshape i\kern-0.25em b}\kern-0.8em\TeX}}}

\copyrightyear{2024} 
\acmYear{2024} 
\setcopyright{acmlicensed}\acmConference[SIGIR '24]{Proceedings of the 47th International ACM SIGIR Conference on Research and Development in Information Retrieval}{July 14--18, 2024}{Washington, DC, USA}
\acmBooktitle{Proceedings of the 47th International ACM SIGIR Conference on Research and Development in Information Retrieval (SIGIR '24), July 14--18, 2024, Washington, DC, USA}
\acmDOI{10.1145/3626772.3657878}
\acmISBN{979-8-4007-0431-4/24/07}

\newcounter{BalanceAtReference}
\newcounter{ReferenceIndexForBalancing}

\makeatletter

\global\@ACM@balancefalse

\def\@balancelastpageonce{%
  \ifnum\value{ReferenceIndexForBalancing}=\value{BalanceAtReference}
    \newpage
  \else
    \relax
  \fi
  \stepcounter{ReferenceIndexForBalancing}
}
\pretocmd{\bibitem}{\@balancelastpageonce}
  {} % on success
  {\@latex@error{Patching \bibitem failed}{\@ehd}}

\usepackage{amsmath}
\usepackage{amsfonts}
\usepackage[linesnumbered,ruled,vlined]{algorithm2e}
\usepackage{graphicx} % DO NOT CHANGE THIS
\usepackage{booktabs}
\usepackage{array}
\usepackage{enumitem}
\usepackage{amsthm}
\usepackage{algpseudocode}
\usepackage[switch]{lineno}
\usepackage[utf8]{inputenc}
\usepackage{eqparbox}
\usepackage[nopar]{lipsum}
\usepackage{multirow}
\usepackage{makecell}
\usepackage{xcolor}
\usepackage{hhline} 
\usepackage{microtype}
\usepackage{diagbox}
\usepackage{longtable} % For longtable in Appendix (Niklas)
\usepackage{adjustbox}
\usepackage{mdframed}
\usepackage{framed,color} 
\usepackage{balance}

\definecolor{shadecolor}{rgb}{.92, .92, .92}
{\endMakeFramed}

\usepackage{relsize}

\usepackage{booktabs,multirow,array}
\newcolumntype{N}{@{}m{0pt}@{}}%a fix for array package

\usepackage[many]{tcolorbox}
\newtcolorbox{fancyquotes}{%
    enhanced jigsaw, 
    breakable,      % allow page breaks
    frame hidden,   % hide the default frame
    left=0.5cm,       % left margin
    right=0.1cm,      % right margin
    overlay={%
        \node [scale=8,
            text=black,
            inner sep=0pt,] at ([xshift=-1cm,yshift=-1cm]frame.north west){}; 
        \node [scale=8,
            text=black,
            inner sep=0pt,] at ([xshift=1cm]frame.south east){};  
            },
                parbox=false,
}

\makeatletter

\newtheorem*{proof*}{Proof}

\usepackage{xcolor}
\usepackage{soul}
\usepackage{colortbl}
\definecolor{benchmark}{HTML}{DAE3F3}
\definecolor{data}{HTML}{E2F0D9}
\definecolor{model}{HTML}{FFE3E3}

\DeclareRobustCommand{\resource}{\textbf{C-Pack}}
\DeclareRobustCommand{\benchmark}{\sethlcolor{benchmark}{\textbf{\hl{C-MTEB}}}}
\DeclareRobustCommand{\data}{\sethlcolor{data}{\textbf{\hl{C-MTP}}}}
\DeclareRobustCommand{\model}{\sethlcolor{model}{\textbf{\hl{BGE}}}}

\newcommand\BibTeX{B\textsc{ib}\TeX}
\usepackage{listings}
\usepackage{color}
\definecolor{codegreen}{rgb}{0.3,0.5,0.0}
\def\@fnsymbol#1{\ensuremath{\ifcase#1\or \dagger\or *\or \ddagger\or
   \mathsection\or \mathparagraph\or \|\or **\or \dagger\dagger
   \or \ddagger\ddagger \else\@ctrerr\fi}}

\newcolumntype{C}[1]{>{\centering\let\newline\\\arraybackslash\hspace{0pt}}m{#1}}

\ExplSyntaxOn
\NewExpandableDocumentCommand { \ValuePlusOne } { m } 
  { \int_eval:n { \int_use:c { c @ #1 } + 1 } }
\NewExpandableDocumentCommand { \Sec } { m } 
  { \fp_eval:n { secd ( #1 ) } }
\NewDocumentCommand { \Rot } { m }
  { 
    \hbox_to_wd:nn { 1 em }
      { 
        \hbox_overlap_right:n 
          { 
            \skip_horizontal:n { \fp_to_dim:n { 7 * cosd (\Angle) } } 
            \rotatebox{\Angle}{#1}
          } 
      } 
  }
\ExplSyntaxOff

\def\Angle{45}
    
\bigskip
\def\Angle{90}

\begin{document}
\title{C-Pack: Packed Resources For General Chinese Embeddings}

\author{Shitao Xiao}
\authornote{These two researchers are co-first authors.}
\email{stxiao@baai.ac.cn}
\affiliation{%
  \institution{Beijing Academy of AI}
  \city{Beijing}
  \country{China}
}

\author{Zheng Liu}
\authornotemark[1]
\authornote{Zheng Liu is the corresponding author}
\email{zhengliu1026@gmail.com}
\affiliation{%
  \institution{Beijing Academy of AI}
  \city{Beijing}
  \country{China}
}

\author{Peitian Zhang}
\email{namespace.pt@gmail.com}
\affiliation{%
  \institution{Renmin University of China}
  \city{Beijing}
  \country{China}
}

\author{Niklas Muennighoff}
\email{n.muennighoff@gmail.com}
\affiliation{%
  \institution{HuggingFace}
  \city{Beijing}
  \country{China}
}

\author{Defu Lian}
\email{liandefu@ustc.edu.cn}
\affiliation{%
  \institution{USTC}
  \city{Hefei}
  \country{China}
}

\author{Jian-Yun Nie}
\email{nie@iro.umontreal.ca}
\affiliation{%
  \institution{University of Montreal}
  \city{Montreal}
  \country{Canada}
}

\renewcommand{\shortauthors}{Shitao Xiao et al.}
%% No italics
%% Use footnote or author note to identify equal contribution and/or contact author info

\begin{CCSXML}
<ccs2012>
   <concept>      
       <concept_id>10002951.10003317.10003338</concept_id>
       <concept_desc>Information systems~Retrieval models and ranking</concept_desc>
       <concept_significance>500</concept_significance>
       </concept>
 </ccs2012>
\end{CCSXML}

\ccsdesc[500]{Information systems~Retrieval models and ranking} 

\begin{abstract}
  We introduce \resource{}, a package of resources that significantly advances the field of general text embeddings for Chinese. \resource{} includes three critical resources.  1) \data{} is a massive training dataset for text embedding, which is  
  based on the curation of vast unlabeled corpora and the integration of high-quality labeled corpora. 2) \benchmark{} is a comprehensive benchmark for Chinese text embeddings covering 6 tasks and 35 datasets. 3) \model{} is a family of embedding models covering multiple sizes. Our models outperform all prior Chinese text embeddings on \benchmark{} by more than +10\% upon the time of the release. We also integrate and optimize the entire suite of training methods for \model{}. Along with our resources on general Chinese embedding, we release our data and models for English text embeddings. The English models also achieve state-of-the-art performance on the MTEB benchmark; meanwhile, our released English data is 2 times larger than the Chinese data. Both Chinese and English datasets are the largest public release of training data for text embeddings. All these resources are made publicly available at { {https://github.com/FlagOpen/FlagEmbedding}}. 
\end{abstract}

\keywords{Text Embeddings, Training Data, Benchmark, Pre-trained Models}

\maketitle

\section{Introduction}
Text embedding is a long-standing topic in natural language processing and information retrieval. By representing texts with latent semantic vectors, text embedding can support various applications, e.g., web search, question answering, and retrieval-augmented language modeling \cite{zhang2022uni, xiao2021matching,lewis2020retrieval,xiao2022distill}. The recent popularity of large language models (LLMs) has made text embeddings even more important. Due to the inherent limitations of LLMs, such as world knowledge and action space, external support via knowledge bases or tool use is necessary. Text embeddings are critical to connect LLMs with these external modules~\cite{borgeaud2022improving,qin2023toolllm}. 

\begin{figure}[t]
\centering
\includegraphics[width=0.95\linewidth]{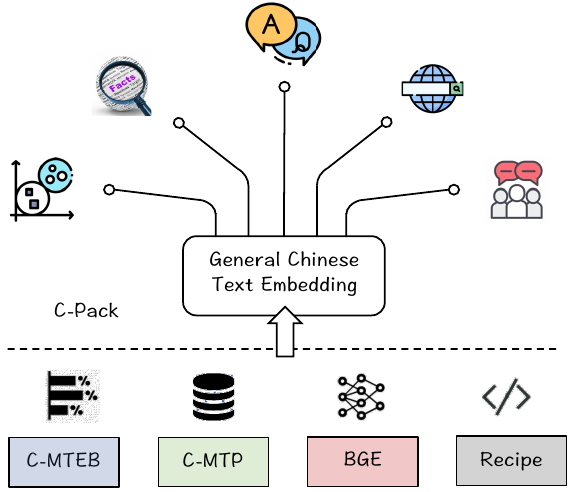}
% \includegraphics[width=0.8\linewidth]{figures/cpack_example.pdf}
% \vspace{-15pt}
% \caption{\textbf{The C-Pack resources to support general Chinese embedding.}}
\caption{C-Pack presents 4 critical resources to support general Chinese embedding: C-MTEB (comprehensive evaluation benchmark), C-MTP (massive training data), BGE (powerful pre-trained models), the entire-suite of training recipe.} 
% \vspace{-10pt}
\label{fig:2}
\end{figure}

The wide variety of application scenarios calls for a single unified embedding model that can handle all kinds of usages (like {retrieval}, {ranking}, {classification}) in any application scenarios (e.g., {question answering}, {language modeling}, {conversation}). However, learning general-purpose text embeddings is much more challenging than task-specific ones. The following factors are critical: 

$\bullet$ \textbf{Data}. The development of general-purpose text embeddings puts forward much higher demands on the training data in terms of \textit{scale}, \textit{diversity}, and \textit{quality}. To achieve high discriminative power for the embeddings, it may take more than hundreds of millions of training instances \cite{izacard2021unsupervised,ni2021large,wang2022text}, which is orders of magnitude greater than typical task-specific datasets, like MS MARCO \cite{nguyen2016ms} and NLI \cite{bowman2015large,williams2017broad}. Besides scale, the training data needs to be collected from a wide range of sources so as to improve the generality across different tasks \cite{izacard2021unsupervised,wang2022text}. Finally, the augmentation of scale and diversity will probably introduce noise. Thus, the collected data must be properly cleaned before being utilized for the training of embeddings \cite{wang2022text}. 

% \footnote{https://huggingface.co/datasets/sentence-transformers/embedding-training-data}

$\bullet$ \textbf{Training}. The training of general-purpose text embeddings depends on two critical elements: a well-suited backbone encoder and an appropriate training recipe. While one can resort to generic pre-trained models like BERT~\cite{devlin2018bert} and T5~\citep{raffel2020exploring}, the quality of text embedding can be substantially improved by pre-training with large-scale unlabeled data \cite{izacard2021unsupervised,wang2022text}. Further, instead of relying on a single algorithm, it takes a compound recipe to train general-purpose text embedding. Particularly, it needs embedding-oriented pre-training to prepare the text encoder \cite{gao2021condenser}, contrastive learning with sophisticated negative sampling to improve the embedding's discriminability \cite{qu2020rocketqa}, and instruction-based fine-tuning \cite{su2022one,asai2022task} to integrate different representation capabilities of text embedding.

$\bullet$ \textbf{Benchmark}. Another pre-requisite condition is the establishment of proper benchmarks, where all needed capabilities of text embeddings can be comprehensively evaluated. BEIR~\cite{thakur2021beir} provides a collection of 18 to evaluate the embedding's general performances on different retrieval tasks, e.g., question answering and fact-checking. Later, MTEB~\cite{muennighoff2022mteb} proposes a more holistic evaluation of embeddings and extends BEIR. It integrates 56 datasets, where all important capabilities of text embeddings, like retrieval, ranking, clustering, etc., can be jointly evaluated. 

Altogether, the development of general-purpose text embedding needs to be made on top of a mixture of driving forces, from data, and encoder models, to training methods and benchmarking. In recent years, continual progresses have been achieved in this field, such as Contriever \cite{izacard2021unsupervised}, E5 \cite{wang2022text}, GTR \cite{ni2021large}, and OpenAI Text Embedding \cite{neelakantan2022text}. {Nevertheless, most of these models are dedicated to the English-centric scenarios. In contrast, there is a shortage of competitive models for general Chinese embedding. What is worse, the development of general Chinese embedding is severely constrained in many aspects: there are neither well-prepared training resources nor suitable benchmarks to evaluate the generality.}\footnote{This situation has been substantially improved since our work. New methods are continually developed on top of the benchmark, data, and models from C-Pack.}

% the progress in Chinese community is limited: there are very few competitive options of public-available models; even worse, there is a severe lack of basic elements, such as benchmarks and training data. 

To address the above challenges, we present a package of resources called \textbf{C-Pack}, which contributes to the development of general Chinese embedding from the following perspectives. 

% \footnote{{Along with our resources on general Chinese embedding, we also publicly release our data and models for English text embeddings. The English models achieve state-of-the-art performance on MTEB benchmark; meanwhile, our released English data is 2 times larger than the Chinese data. Interested readers may check our Github repo for more details.}}

$\bullet$ \benchmark{} (Chinese Massive Text Embedding Benchmark). The benchmark is established as a Chinese extension of MTEB.\footnote{https://huggingface.co/spaces/mteb/leaderboard} \benchmark{} collects 35 public-available datasets belonging to 6 types of tasks. We set up the unified testing protocols so that different embeddings can be evaluated on fair ground. We also develop the evaluation pipeline which significantly makes ease for the evaluation process. Thanks to the scale and diversity of \benchmark{}, all major capabilities of Chinese embeddings can be reliably measured, making it the most suitable benchmark to evaluate the generality of Chinese text embedding.

% It is established as an Chinese extension of MTEB\footnote{https://huggingface.co/spaces/mteb/leaderboard}, which covers 6 types of evaluation tasks and comprises 35 datasets, making it the most suitable benchmark to evaluate the generality of Chinese text embeddings. 

$\bullet$ \data{} (Chinese Massive Text Pairs). We create a massive training dataset of 100M text pairs. The majority of our dataset is curated from the massive web corpora, such as Baike (Wikipedia-style webs in Chinese), Zhihu (a major Chinese social media), major News Websites in Chinese. We extract the semantically related text pairs leveraging the rich-structured information within the data, such as title-to-document, subtitle-to-passage, question-to-answer, question-to-similar-question, etc. The extracted data is further cleaned for the massive weakly supervised training of the text embeddings. We also integrate diverse labeled datasets, which presents high-quality supervision signals for the final refinement of text embeddings. Besides, considering that there is no public available dataset for general English text embeddings either, we curate another massive dataset for English with the same method, which consists of 200M text pairs.

% which integrates both labeled data and unlabeled data curated from Wudao~\cite{yuan2021wudaocorpora}, one of the largest corpora for pre-training Chinese language models. \data{} is not only large and diverse but also cleaned to ensure the data quality.  

$\bullet$ \model{} (BAAI General Embeddings). We provide a family of well-trained models for Chinese general text embeddings. There are three optional model sizes: small (24M), base (102M), and large (326M), which present users with the flexibility to trade off efficiency and effectiveness. Our models make a big leap forward in generality: \model{} outperforms all previously Chinese text embedding models on all aspects of \benchmark{} by large margins. Besides being directly applicable, \model{} can also be fine-tuned with additional data for better downstream performances. The releasing of these powerful models substantially contributes to critical applications, such as search, question answering, and retrieval-augmented generation. 

$\bullet$ \textbf{Training Recipe}. Accompanying our resources, we integrate and optimize training methods to build general-purpose text embeddings, including the pre-training of an embedding-oriented text encoder, general-purpose contrastive learning, and task-specific fine-tuning. The release of the training recipe will help the community to reproduce the state-of-the-art methods and make continuous progress on top of them. 

% Together with our released data, people may easily reproduce and continue to improve the current models. 

% We also integrate and optimize the training methods towards general-purpose text embeddings. Taking account of both effectiveness and the simplicity for massive-scale training, we pipeline the following training recipe: the pre-training with RetroMAE \cite{liu2022retromae}, the contrastive learning with DPR \cite{karpukhin2020dense}, and the multi-task learning with Instructor \cite{su2022one} and LoRA \cite{hu2021lora}. Together with our released data, people may easily reproduce and continue to improve the current models. 

Our project enjoys a widespread popularity in technique communities, like HuggingFace and Github. Remarkably, according to the up-to-date statics (2024-04), BGE model series have received more than \textbf{20 million downloads} from HuggingFace since its release on 2023-08, making it one of the most popular embedding models in the world. 
They have been integrated by the major RAG and text-embedding frameworks in the world, such as Langchain\footnote{\scriptsize https://python.langchain.com/docs/integrations/text\_embedding/bge\_huggingface}, LLamaIndex\footnote{\scriptsize https://docs.llamaindex.ai/en/stable/examples/embeddings/huggingface.html}, and Huggingface\footnote{\scriptsize https://huggingface.co/docs/text-embeddings-inference/index}. 
The code-base also receives nearly \textbf{5,000 stars} on GitHub. So far, there have been over 100 submissions on C-MTEB, and it is widely recognized as the \textbf{most popular and authoritative benchmark} for Chinese text embeddings. 
It's worth noting that our project is still fast growing, with new resources continually created and released to the public. In summary, C-Pack provides a go-to option for the \textbf{development}, \textbf{evaluation}, and \textbf{application} of general-purpose Chinese text embedding, which establishes a solid foundation for the advancement of this field. 

The remaining part of this paper is organized as follows. We discuss the related works about general text embeddings in Section \ref{sec:relate}. We present a detailed introduction about C-Pack resources in Section \ref{sec:cpack}. We make comprehensive empirical analysis for the value of C-Pack in Section \ref{sec:exp}. Finally, we conclude our work in Section \ref{sec:conclude}.

% Within just a few months, The resources have received millions of downloads from the community users for both development and application purposes. MTEB has become an authoritative benchmark for Chinese embedding. BGE has been integrated by the major RAG and text-embedding frameworks in the world, such as Langchain\footnote{\scriptsize https://python.langchain.com/docs/integrations/text\_embedding/bge\_huggingface}, LLamaIndex\footnote{\scriptsize https://docs.llamaindex.ai/en/stable/examples/embeddings/huggingface.html}, and Huggingface\footnote{\scriptsize https://huggingface.co/docs/text-embeddings-inference/index}. It's worth noting that our project is still fast growing, with new resources continually created and released to the public. In summary, C-Pack provides a go-to option for the \textbf{development}, \textbf{evaluation}, and \textbf{application} of general-purpose Chinese text embedding, which establishes a solid foundation for the advancement of this field. 

\begin{figure*}[t]
\centering
\includegraphics[width=1.0\linewidth]{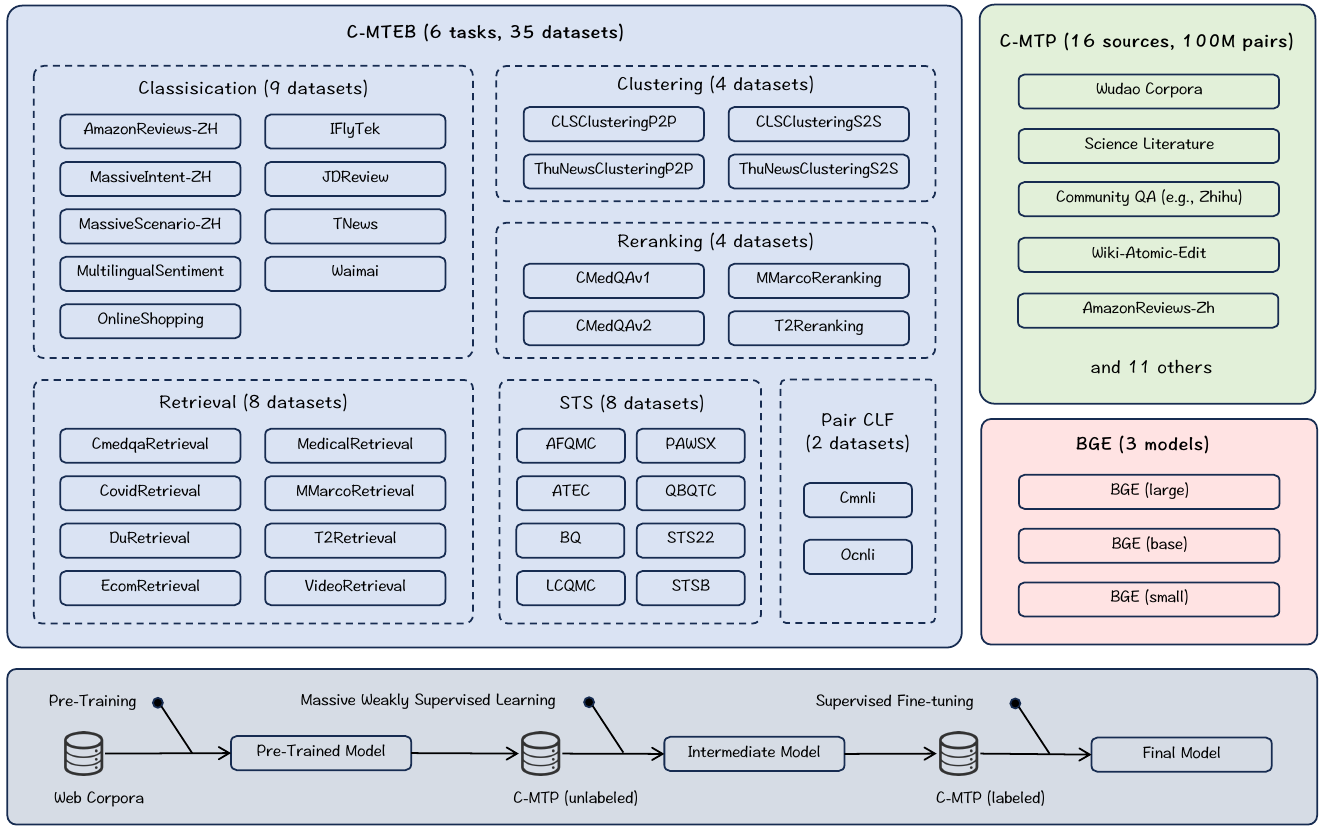}
% \vspace{-15pt}
%\caption{There are three critical resources offered by C-Pack: C-MTEB, C-MTP, C-TEM. The training recipe is shown as below, where three steps are consecutively performed on Wudao corpora, C-MTP (unlabeled) \& (labeled).}
%%% Suggestion from Niklas %%%
% \caption{\textbf{Overview of C-MTE, Chinese Massive Text Embedding resources.} \benchmark{} is a benchmark for Chinese text embeddings. \data{} is a large-scale Chinese embedding training dataset. \model{} are state-of-the-art Chinese embedding models. The training recipe is shown at the bottom.}
\caption{\textbf{Overview of C-Pack.} \benchmark{} is a benchmark for Chinese text embeddings. \data{} is a large-scale Chinese embedding training dataset. \model{} are state-of-the-art Chinese embedding models. The training recipe is shown at the bottom.}
%%% End Suggestion %%%
\vspace{-5pt}
\label{fig:1}
\end{figure*}

\section{Related Work}\label{sec:relate}
% text embedding: key techniques
% general purpose text embedding 
The importance of general text embedding is widely recognized, not only for its wide usage in typical applications, like web search and question answering \cite{xiao2022progressively,karpukhin2020dense} but also due to its fundamental role in augmenting large language models \cite{lewis2020retrieval,guu2020retrieval,borgeaud2022improving,izacard2022few,shi2023replug}. Compared with the conventional task-specific methods, the general text embedding needs to be extensively applicable in different scenarios. In recent years, there has been a continual effort in this field, where a series of well-known works are proposed, like Contriever \cite{izacard2021unsupervised}, GTR \cite{ni2021large}, sentence-T5 \cite{ni2021sentence}, Sentence-Transformer \cite{reimers2019sentence}, E5 \cite{wang2022simlm}, OpenAI text embedding \cite{neelakantan2022text}, etc. Although it remains an open problem, recent studies highlight the following important factors. 

$\bullet$ Firstly, the training data is desired to be large-scale and diversified, from which the embedding model can learn to recognize different kinds of semantic relationships \cite{izacard2021unsupervised,ni2021large,wang2022text,neelakantan2022text}. It usually calls for the comprehensive and elaborate data curation from web corpora, such as online encyclopedia, QA platforms, news websites, and social media communities. Despite similar efforts made by the previous works, few of the curated datasets are made public available before C-Pack. 

$\bullet$ Secondly, the embedding model must be scaled up in terms of both training and model size, as scaled-up text encoders are more generalizable across different application scenarios \cite{muennighoff2022sgpt,ni2021large,ni2021sentence}. Such an observation is in line with the conclusion for the importance of scaling LLMs~\citep{hoffmann2022training,rae2021scaling,brown2020language,chowdhery2022palm,srivastava2022beyond,eval-harness,li2023starcoder,allal2023santacoder,muennighoff2023scaling}. Although large-scale training is affordable for many industrial organizations and companies, it remains a huge burden for the community users. The public release of BGE saves such a cost, benefiting both direct applications of text embeddings and further improvements for specific scenarios. 

$\bullet$ Thirdly, the training recipe must be optimized through pre-training \cite{liu2022retromae,wang2022simlm}, negative sampling \cite{izacard2021unsupervised,wang2022simlm}, and multi-task fine-tuning \cite{su2022one,asai2022task,sanh2021multitask,wei2021finetuned,muennighoff2022crosslingual,muennighoff2023octopack,chung2022scaling}. In C-Pack, these operations are integrated, optimized, and pipelined, which significantly facilitates people's reproduction and continual fine-tuning of BGE. 

$\bullet$ Aside from the above factors, it is also critical to establish proper benchmarks to evaluate the generality of text embeddings. Unlike previous task-specific evaluations, like MSMARCO \cite{nguyen2016ms}, SentEval \cite{conneau2018senteval}, it is needed to substantially augment the benchmarks so as to evaluate the embedding's performance for a wide variety of tasks. One representative work is made by BEIR~\cite{thakur2021beir,kamalloo2023resources}, where the embeddings can be evaluated across different retrieval tasks. It is later extended by MTEB \cite{muennighoff2022mteb}, where all major aspects of text embeddings can be comprehensively evaluated. However, no such works were done for the Chinese community before. By introducing C-MTEB, this limitation has been substantially conquered.  

Given the above analysis, it can be concluded that the general text embedding is highly resource-dependent, which calls for a wide range of elements, such as datasets, models, and benchmarks. Thus, the creation and public release of the corresponding resources in C-Pack is crucially important.

\section{C-Pack}\label{sec:cpack}
In this section, we first introduce the resources in C-Pack: the benchmark \benchmark{}, the training data \data{}, and the model class \model{}. Then, we discuss the training recipe, which enables us to train the state-of-the-art models for general Chinese embedding based on the offered resources. 

\subsection{Benchmark: C-MTEB}\label{sec:cmteb}
% \subsection{Benchmark: \benchmark{}}\label{sec:cmteb}
% 6 class of tasks, 35 datasets, analogue to MTEB 
% Introduction of the tasks
% Specification of the datasets
% incorporated into MTEB, as a Chinese specific branck
% dataset stats
% experiment
% observations 

\benchmark{} is established for the comprehensive evaluation of the generality of Chinese embeddings (\autoref{fig:1}). In the past few years, the community has put forward many datasets for text representation and language understanding tasks in Chinese, such as CMNLI \cite{xu2020clue}, DuReader \cite{he2017dureader}, T$^2$Ranking \cite{xie2023t2ranking}. However, these datasets are independently curated, lacking a fair and shared ground to comprehensively evaluate the general capability of text embeddings. Therefore, we create \benchmark{}, where the following important efforts are made: 1) the comprehensive collection of datasets which can be either directly utilized or repurposed for the evaluation of text embeddings, 2) the categorization of the datasets into different capability attributes of text embeddings, e.g., retrieval, similarity analysis, classification, etc., 3) the standardization of the evaluation protocols, 4) the establishment of evaluation pipelines. 

In particular, we collect a total of 35 public datasets related to the evaluation of Chinese text embeddings (Briefed as Figure \ref{fig:2}. Detailed specifications are presented in the Github Repository\footnote{https://github.com/FlagOpen/FlagEmbedding/tree/master/C\_MTEB}). The collected datasets are categorized based on the embedding's capability they may evaluate. There are 6 groups of evaluation tasks: retrieval, re-ranking, STS (semantic textual similarity), classification, pair classification, and clustering, which cover the main interesting aspects of Chinese text embeddings. Note that there are multiple datasets for each category. The datasets of the same category are collected from different domains and complementary to each other, therefore ensuring the corresponding capability to be fully evaluated.

The nature of each evaluation task and the evaluation metric are briefly introduced as follows.  

$\bullet$ \textbf{Retrieval}. The retrieval task is presented with the test queries and a large corpus. For each query, it finds the Top-$k$ similar documents within the corpus. The retrieval quality can be measured by ranking and recall metrics at different cut-offs. In this work, we use the setting from BEIR \cite{thakur2021beir}, using NDCG@10 as the main metric. 

$\bullet$ \textbf{Re-ranking}. The re-ranking task is presented with test queries and their candidate documents (1 positive plus N negative documents). For each query, it re-ranks the 
documents based on the embedding similarity. The MAP score is used as the main metric. 

$\bullet$ \textbf{STS} (Semantic Textual Similarity). The STS~\citep{agirre2012semeval,agirre2013sem,agirre2014semeval,agirre2015semeval,agirre2016semeval} task is to measure the correlation of two sentences based on their embedding similarity. Following the original setting in Sentence-BERT \cite{reimers2019sentence}, the Spearman's correlation is computed with the given label, whose result is used as the main metric. 

% $\bullet$ \textbf{Classification}. The classification task re-uses the logistic regression classifier from MTEB \cite{muennighoff2022mteb}, where the provided label is predicted based on the input embedding. The average precision is used as the main metric. 

$\bullet$ \textbf{Classification}. The classification task re-uses the logistic regression classifier from MTEB \cite{muennighoff2022mteb}, where the average precision is used as the main metric. 

$\bullet$ \textbf{Pair-classification}. This task deals with a pair of input sentences, whose relationship is presented by a binarized label. The relationship is predicted by embedding similarity, where the average precision is used as the main metric. 

$\bullet$ \textbf{Clustering}. The clustering task is to group sentences into meaningful clusters. Following the original setting in MTEB \cite{muennighoff2022mteb}, it uses the mini-batch k-means method for the evaluation, with batch size equal to 32 and k equal to the number of labels within the mini-batch. The V-measure score is used as the main metric. 

\begin{figure}[t]
    \centering
    \includegraphics[width=1.0\linewidth]{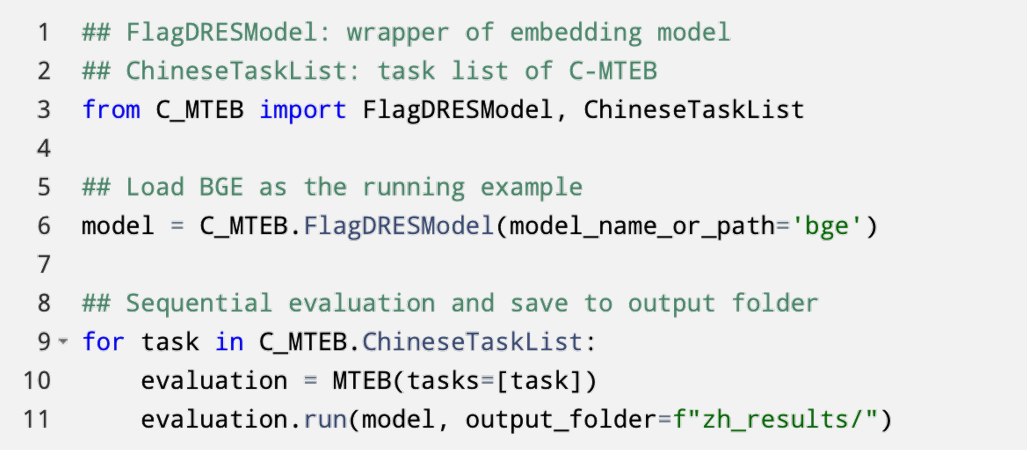} 
    % \vspace{-5pt} 
    \caption{The evaluation pipeline of C-MTEB.} 
    \label{fig:cmteb}
    % \vspace{-5pt}
\end{figure}

Finally, the embedding's capability on each task is measured by the average performance of all datasets for that task. The embedding's overall generality is measured by the average performance of all datasets in \benchmark{}. We set up the standardized and compact pipeline for all tasks where different embedding models can be evaluated on a fair basis (Figure \ref{fig:cmteb}). FlagDRESModel is the wapper for the custmoized embedding model, which implements the encoding method for query and document. ChineseTaskList is the task list of C-MTEB. The evaluation result is written to the output folder, which can be directly submitted to the C-MTEB leaderboard\footnote{https://huggingface.co/spaces/mteb/leaderboard}. 

% \blue{We also establish the evaluation pipeline for C-MTEB. Figure~\ref{fig:evaluation} shows a simple example. Users can evaluate the performance of embedding model on C-MTEB with just a few lines of code, and all results will be stored in the output folder. The saved results can be used to submit to the online MTEB leaderboard on HuggingFace platform.} 

% \begin{algorithm}
% \begin{lstlisting}
% model = C_MTEB.FlagDRESModel(model_name_or_path='bge')
% for task in C_MTEB.ChineseTaskList:
%     evaluation = MTEB(tasks=[task])
%     evaluation.run(model, output_folder=f"zh_results/")
% \end{lstlisting}
% \caption{Evaluation for C-MTEB}\label{alg:vcs}
% \end{algorithm}

% \begin{figure}[!h]
% \centering
% \begin{minipage}{0.96\linewidth}
% \begin{lstlisting}[language=Python]
% from C_MTEB import FlagDRESModel, ChineseTaskList

% model = FlagDRESModel('BAAI/bge-small-zh')
% for task in ChineseTaskList:
%     evaluation = MTEB(tasks=[task])
%     evaluation.run(model, output_folder=f"results/")
% \end{lstlisting}
% \end{minipage} 
% \vspace{-20pt}
% \caption{An example of evaluation for C-MTEB}
% % \vspace{-5pt}
% \label{fig:evaluation} 
% \end{figure}

% The evaluation of C-MTEB is integrated into the well-etablished pipeline of MTEB\footnote{https://github.com/embeddings-benchmark/mteb/ blob/main/mteb/cmd.py}.

% \begin{mdframed}[
%       backgroundcolor=shadecolor,
%       leftmargin=3pt,
%       innerleftmargin=7pt,
%       innertopmargin=5pt,
%       skipabove=5pt,
%       skipbelow=5pt
%     ]
% \footnotesize{
% Eval = MTEB(tasks=args.task, task\_langs=args.lang)
% }
% \end{mdframed} 

\subsection{Training Data: C-MTP} 
% \subsection{Training Data: \data{}}
We curate the largest dataset \data{} for the training of general Chinese embedding. The paired texts constitute the data foundation for the training of text embedding, e.g., a question and its answer, two paraphrase sentences, or two documents on the same topic. To ensure the generality of the text embedding, the paired texts need to be both large-scale and diversified. Therefore, \data{} is collected from two sources. The majority of the data is based on the curation of massive unlabeled data, \textit{a.k.a.} \data{} \textbf{(unlabeled)}, which presents 100 millions of paired texts. Meanwhile, a small portion is from the comprehensive integration of high-quality labeled data, \textit{a.k.a.} \data{} \textbf{(labeled)}, which leads to about 1 million paired texts. The data collection process is briefly introduced as follows. 

\begin{table}[t]
    \centering
    % \footnotesize
    \caption{\textbf{Composition of \data{}}}
    \begin{tabular}{|p{1.2cm}|C{2.8cm}|C{2.8cm}|}
    % \ChangeRT{1pt} 
    \hline
    dataset & \textbf{\data{} (unlabeled)} & \textbf{\data{}  ~~~~ (labeled)}  \\
    \hline
    source  & Wudao, Zhihu, Baike, CSL, XLSUM-Zh, Amazon-Review-Zh, CMRC, etc. & T$^2$-Ranking, mMARCO-Zh, DuReader, NLI-Zh, etc. \\
    \hline
    size & 100$M$ & 838$K$ \\
    \hline
    % \ChangeRT{1pt}
    \end{tabular}
    % \vspace{-5pt}
    % \caption{\textbf{Composition of \data{}}}
    % \vspace{-5pt}
    \label{tab:2}
\end{table}

\begin{figure}[ht!]
    \centering
    % \caption{Creation of C-MTP.}   
    \includegraphics[width=1.0\linewidth]{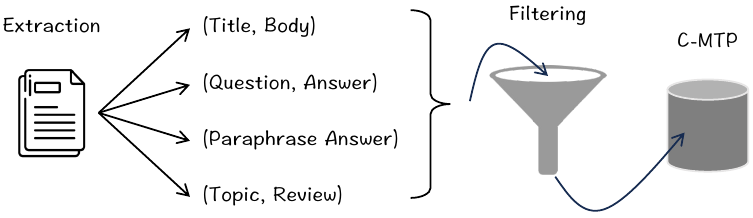} 
    % \vspace{-5pt} 
    \caption{Creation of C-MTP.} 
    \label{fig:cmtp}
    % \vspace{-5pt}
\end{figure}

% the curation of massive unlabeled data, \textit{a.k.a.} \data{} \textbf{(unlabeled)}; and the comprehensive collection of labeled data, \textit{a.k.a.} \data{} \textbf{(labeled)}.  The data collection process is briefly introduced as follows. 

$\bullet$ \textbf{\data{} (unlabeled)}. We look for a wide variety of corpora, where we can extract rich-semantic paired structures from the plain text, e.g., paraphrases, title-body. Our primary source of data comes from open web corpora. The most representative one is the Wudao corpus \cite{yuan2021wudaocorpora}, which is the largest dataset of well-formatted articles for pre-training Chinese language models. For each its article, we extract structures, like (title, body), (sub-title, passage), to form a text pair. In addition, we collect data from other web content like Zhihu, Baike, news websites, which complement other forms of text pairs, especially (question, answer), (paraphrase titles), (paraphrase answers), etc. Aside from the open web content, we also explore other public Chinese data for more diverse text pairs, such as CSL (scientific literature), Amazon-Review-Zh (topic and its reviews), Wiki Atomic Edits (paraphrases), CMRC (machine reading comprehension), XLSUM-Zh (summarization), etc. 

% we extract (title, passage) to form a text pair. Following the same recipe, we also collect such text pairs from other similar web content like Zhihu, Baike, news websites, etc. Aside from the open web content, we also explore other public Chinese datasets to extract text pairs, such as CSL (scientific literature), Amazon-Review-Zh (reviews), Wiki Atomic Edits (paraphrases), CMRC (machine reading comprehension), XLSUM-Zh (summarization), etc. The paired structures are obvious in these datasets, which are directly extracted for the augmentation of \data{} \textbf{(unlabeled)}. 

The text pairs curated from the web and other public sources are not guaranteed to be closely related. Therefore, data quality can be a major concern. In our work, we make use of a compound \textbf{data cleaning} strategy to refine the raw data. Firstly, the whole data undergoes general filtering, which removes non-textual, duplicated, and malicous content. Secondly, the data is further processed by semantic filtering so as to ensure the text pairs are semantically related. In our work, we make use a third-party model: Text2Vec-Chinese\footnote{https://huggingface.co/GanymedeNil} to score the strength of relation for each text pair. We empirically choose a threshold of 0.43, and drop the samples whose scores are below the threshold. With such an operation, there are 100 million text pairs filtered from the unlabeled corpora. Despite the simplicity, we find that it effectively removes the irrelevant text pairs when manually reviewing samples and leads to strong empirical performances for the models trained on \data{} \textbf{(unlabeled)}.

% we use a simple strategy to filter the data before adding it to \data{} \textbf{(unlabeled)}. Particularly, we use a third-party model: Text2Vec-Chinese\footnote{https://huggingface.co/GanymedeNil} to score the strength of relation for each text pair. We empirically choose a threshold of 0.43, and drop the samples whose scores are below the threshold. With such an operation, there are 100 million text pairs filtered from the unlabeled corpora. Despite the simplicity, we find that it effectively removes the irrelevant text pairs when manually reviewing samples and leads to strong empirical performances for the models trained on \data{} \textbf{(unlabeled)}. 

$\bullet$ \textbf{\data{} (labeled)}. The labeled data is collected to further enhance to the training data. In our work, we integrate and re-purpose a diverse group of datasets, which covers different capabilities of the text embedding, like retrieval, ranking, similarity comparison, etc. Particularly, the following labeled datasets are included, T$^2$-Ranking \cite{xie2023t2ranking}, DuReader \cite{he2017dureader,qiu2022dureader_retrieval}, mMARCO \cite{bonifacio2021mmarco}, CMedQA-v2\cite{cmedqa}, multi-cpr\cite{long2022multi}, NLI-Zh\footnote{https://huggingface.co/datasets/shibing624/nli\_zh}, cmnli\cite{xu2020clue} and ocnli\cite{xu2020clue}. There are 838,465 paired texts in total, which contains diverse question-answering and paraphrasing patterns. Although it is much smaller than \data{} \textbf{(unlabeled)}, most of the data is curated from human annotation, thus ensuring a high credibility of relevance.

% The following labeled datasets are collected for \data{} \textbf{(labeled)} due to their quality and diversity: T$^2$-Ranking \cite{xie2023t2ranking}, DuReader \cite{he2017dureader,qiu2022dureader_retrieval}, mMARCO \cite{bonifacio2021mmarco}, CMedQA-v2\cite{cmedqa}, multi-cpr\cite{long2022multi}, NLI-Zh\footnote{https://huggingface.co/datasets/shibing624/nli\_zh}, cmnli\cite{xu2020clue} and ocnli\cite{xu2020clue}. There are 838,465 paired texts in total. Although it is much smaller than \data{} \textbf{(unlabeled)}, most of the data is curated from human annotation, thus ensuring a high credibility of relevance. Besides, \data{} \textbf{(labeled)} also fully covers different capabilities of the text embedding, like retrieval, ranking, similarity comparison, etc., which helps to improve the embedding model's generality after fine-tuning. 

Given the differences in scale and quality, \data{} \textbf{(unlabeled)} and \data{} \textbf{(labeled)} are applied to different training stages, which jointly result in a strong performance for the embedding model. Detailed analysis will be made in our training recipe. 

% \begin{figure}[t]
%     \centering
%     \includegraphics[width=0.95\linewidth]{figures/cmtp.pdf} 
%     \vspace{-5pt} 
%     \caption{Creation of C-MTP.} 
%     \label{fig:cmtp}
%     % \vspace{-5pt}
% \end{figure}

% \begin{table}[ht!]
%     \centering
%     \footnotesize
%     \begin{tabular}{|p{1.0cm}|C{2.5cm}|C{2.5cm}|}
%     % \ChangeRT{1pt} 
%     \hline
%     dataset & \textbf{\data{} (unlabeled)} & \textbf{\data{} ~~~~ (labeled)}  \\
%     \hline
%     source  & Wudao, Zhihu, Baike, CSL, XLSUM-Zh, Amazon-Review-Zh, CMRC, etc. & T$^2$-Ranking, mMARCO-Zh, DuReader, NLI-Zh, etc. \\
%     \hline
%     size & 100$M$ & 838$K$ \\
%     \hline
%     % \ChangeRT{1pt}
%     \end{tabular}
%     % \vspace{-5pt}
%     \caption{\textbf{Composition of \data{}}}
%     % \vspace{-10pt}
%     \label{tab:2}
% \end{table} 

\subsection{Model Class: BGE}
% \subsection{Model Class: \model{}}
% what kind of model we release (the final models, the intermediate ckpt)
% the values: high-precision, trade-off between precision and efficiency, directly usable and finetunable 

Even with the full package of training data, it is still challenging to learn general text embeddings due to the expensive training process. In our work, we provide a comprehensive class of well-trained embedding models for the community. Our models are based on the BERT-like architecture \cite{devlin2018bert}, which go through three-stage of training (to be discussed in the next section). 
% Our models take a BERT-like architecture, where the last layer's hidden state of the special token [CLS] is trained to work as the embedding.  
% \begin{equation}\label{eq:1}
%     \mathbf{e}_ = \mathrm{BERT}()
% \end{equation}
There are three available scales: large (with 326M parameters), base (with 102M parameters), and small (with 24M parameters). The large-scale model achieves the highest general representation performances, leading the current public-available models by a considerable margin. The small-scale model is also empirically competitive compared with the public-available models and other model options in \model{}; besides, it is way faster and lighter, making it suitable to handle massive knowledge bases and high-throughput applications. Thanks to the comprehensive coverage of different model sizes, people are presented with the flexibility to trade off running efficiency and representation quality based on their own needs. 

As introduced, the models within \model{} have been well-trained and achieve a strong generality for a wide variety of tasks. 
% Meanwhile, they can also be further fine-tuned if 
Meanwhile, they also establish a strong foundation for further fine-tuning. The fine-tuning recipe is well formulated in our training recipe, where all people's need is the preparation of fine-tuning data. 
% 1) the embeddings are applied for a specific scenario, 2) the training data is presented for the application scenario. 
It is empirically verified that the fine-tuned model may bring forth a much better performance for its application, compared with its original model in \model{}, and the fine-tuned models from other general pre-trained encoders, like BERT. In other words, \model{} not only presents people with direct usage embeddings but also works as a foundation where people may develop more powerful embeddings.

\begin{table*}[t]
    \centering
    % \small
    % \footnotesize
    % \begin{tabular}{|p{1.0cm}|C{2.5cm}|C{2.5cm}|}
    \caption{\textbf{Performance of various models on \benchmark{}.}}
    \begin{tabular}{|p{2.8cm}|C{1.2cm}|C{1.2cm}|C{1.2cm}|C{1.2cm}|C{1.2cm}|C{1.2cm}|C{1.2cm}|C{1.2cm}|}
    % \ChangeRT{1pt} 
    \hline
    model & Dim & Retrieval & STS & Pair CLF & CLF & Re-rank & Cluster & Average  \\
    \hline
    Text2Vec (base) & 768 & 38.79 & 43.41 & 67.41 & 62.19 & 49.45 & 37.66 & 48.59 \\
    Text2Vec (large) & 1024 & 41.94 & 44.97 & 70.86 & 60.66 & 49.16 & 30.02 & 48.56 \\
    Luotuo (large) & 1024 & 44.40 & 42.79 & 66.62 & 61.0 & 49.25 & 44.39 & 50.12 \\
    M3E (base) & 768 & 56.91 & 50.47 & 63.99 & 67.52 & 59.34 & 47.68 & 57.79 \\
    M3E (large) & 1024 & 54.75 & 50.42 & 64.30 & 68.20 & 59.66 & \textbf{48.88} & 57.66 \\
    Multi. E5 (base) & 768 & 61.63 & 46.49 & 67.07 & 65.35 & 54.35 & 40.68 & 56.21 \\
    Multi. E5 (large) & 1024 & 63.66 & 48.44 & 69.89 & 67.34 & 56.00 & 48.23 & 58.84 \\
    OpenAI-Ada-002 & 1536 & 52.00 & 43.35 & 69.56 & 64.31 & 54.28 & 45.68 & 53.02 \\ 
    \hline
    BGE (small) & 512 & 63.07 & 49.45 & 70.35 & 63.64 & 61.48 & 45.09 & 58.28 \\
    BGE (base)  & 768 & 69.53 & 54.12 & 77.50 & 67.07 & 64.91 & 47.63 & 62.80 \\
    BGE (large) & 1024 & \textbf{71.53} & \textbf{54.98} & \textbf{78.94} & \textbf{68.32} & \textbf{65.11} & {48.39} & \textbf{63.96} \\
    \hline
    % \ChangeRT{1pt}
    \end{tabular}
    % \vspace{-5pt}
    % \caption{\textbf{Performance of various models on \benchmark{}.}}
    % \vspace{-10pt}
    \label{tab:3}
\end{table*}

\subsection{Training Recipe}
% three stages: pre-train, contrastive learning (unlabeled), supervised learning (intruction tuning) 

The training recipe of \model{} is completely released to the public along with C-Pack (\autoref{fig:1}). Our training recipe has three main components: \textbf{1)} pre-training with plain texts, \textbf{2)} contrastive learning with \data{} \textbf{(unlabeled)}, and \textbf{3)} multi-task learning with \data{} \textbf{(labeled)}. As introduced, the public release of training recipe will not only help with the reproduction, but also benefit the improvement of BGE with continual training and fine-tuning. 

% , whose specifications are made as follows. 

$\bullet$ \textbf{Pre-Training}. Our model is pre-trained on massive plain texts through a tailored algorithm in order to better support the embedding task. Particularly, we make use of the Wudao corpora \cite{yuan2021wudaocorpora}, which is a huge and high-quality dataset for Chinese language model pre-training. We leverage the MAE-style approach presented in RetroMAE \cite{liu2022retromae,xiao2023retromae}, which is simple but highly effective. The polluted text is encoded into its embedding, from which the clean text is recovered on top of a light-weight decoder:
\begin{equation*}\label{eq:2}
    \min. \sum_{x \in X} -\log\mathrm{Dec}(x|\mathbf{e}_{\tilde{X}}), ~ \mathbf{e}_{\tilde{X}} \leftarrow \mathrm{Enc}(\tilde{X}).
\end{equation*}
($\mathrm{Enc}$, $\mathrm{Dec}$ indicate the encoding and decoding operations, $X$, $\tilde{X}$ indicate the clean and polluted text.) 

$\bullet$ \textbf{General purpose fine-tuning}. The pre-trained model is fine-tuned on \data{} \textbf{(unlabeled)} via contrastive learning, where it is learned to discriminate the paired texts from their negative samples:
\begin{equation*}
    \min. \sum_{(p,q)} - \log \frac{e^{\langle \mathbf{e}_p, \mathbf{e}_{q}\rangle / \tau}}{e^{\langle \mathbf{e}_p, \mathbf{e}_{q}\rangle  / \tau} + \sum_{Q'} e^{\langle \mathbf{e}_p, \mathbf{e}_{q'}\rangle / \tau} }.
\end{equation*}
($p$ and $q$ are the paired texts, $q'\in Q'$ is a negative sample, $\tau$ is the temperature). One critical factor of contrastive learning is the negative samples. Instead of mining hard negative samples on purpose, we purely rely on in-batch negative samples \cite{karpukhin2020dense} and resort to a big batch size (as large as 19,200) to improve the discriminativeness of the embedding. 

$\bullet$ \textbf{Task-specific fine-tuning}. The embedding model is further fine-tuned with \data{} \textbf{(labeled)}. The labeled datasets are smaller but of higher quality. However, the contained tasks are of different types, whose impacts can be mutually contradicted. In this place, we apply two strategies to mitigate this problem. On one hand, we leverage instruction-based fine-tuning \cite{su2022one,asai2022task}, where the input is differentiated to help the model accommodate different tasks. 
% For each text pair in retrieval and QA tasks (T$^2$-Ranking, mMARCO-zh, and DuReader), an instruction $I_{t}$: "为这个句子生成表示以用于检索相关文章：" is attached to the query side: $q' \leftarrow q + I_{t}$. 
For each text pair ($p$, $q$), a task specific instruction $I_{t}$ is attached to the query side: $q' \leftarrow q + I_{t}$. The instruction is a verbal prompt, which specifies the nature of the task, e.g., ``\textit{search relevant passages for the query}''. 
On the other hand, the negative sampling is updated: in addition to the in-batch negative samples, one hard negative sample $q'$ is mined for each text pair ($p$, $q$). The hard negative sample is mined from the task's original corpus, following the ANN-style sampling strategy in \cite{xiong2020approximate}.

\section{Experiments}\label{sec:exp}
% general performance
% impact from different stage
% batch size
% instruction
% pre-training

% detailed performance 

In this section, we conduct experimental studies for the exploration of following problems. \textbf{P1}. The extensive evaluation of different Chinese text embeddings on \benchmark{}. \textbf{P2}. The empirical verification of the text embeddings by \model{}. \textbf{P3}. The exploration of the practical value brought by \data{}. \textbf{P4}. The exploration of the impacts introduced by the training recipe. We consider the following popular Chinese text embedding models as the baselines for our experiments: Text2Vec-Chinese\footnote{https://huggingface.co/shibing624} base and large; {Luotuo}\footnote{https://huggingface.co/silk-road/luotuo-bert-medium}; M3E\footnote{https://huggingface.co/moka-ai} base and large; multilingual E5~\cite{wang2022text} and OpenAI text embedding ada 002\footnote{https://platform.openai.com/docs/guides/embeddings}. The main metric presented in Section \ref{sec:cmteb} is reported for each evaluation task in \benchmark{}.

\begin{table*}[t]
    \centering
    % \small
    % \footnotesize
    % \begin{tabular}{|p{1.0cm}|C{2.5cm}|C{2.5cm}|}
    \caption{\textbf{Ablation of the training data, \data{}, and the training recipe.}}
    \begin{tabular}{|p{2.8cm}|C{1.1cm}|C{1.2cm}|C{1.2cm}|C{1.2cm}|C{1.2cm}|C{1.2cm}|C{1.2cm}|C{1.2cm}|}
    % \ChangeRT{1pt} 
    \hline
    model & Dim & Retrieval & STS & Pair CLF & CLF & Re-rank & Cluster & Average  \\
    \hline
    %Text2Vec (large) & 1024 & 41.94 & 44.97 & 70.86 & 60.66 & 49.16 & 30.02 & 48.56 \\
    M3E (large)      & 1024 & 54.75 & 50.42 & 64.30 & 68.20 & 59.66 & 48.88 & 57.66 \\
    OpenAI-Ada-002   & 1536 & 52.00 & 43.35 & 69.56 & 64.31 & 54.28 & 45.68 & 53.02 \\
    \hline 
    % bz:256           & 1024 & 57.25 & 46.16 & 62.02 & 65.71 & 58.59 & 49.52 & 56.43 \\
    % bz:2048          & 1024 & 60.96 & 46.60 & 61.91 & 67.42 & 59.98 & 49.04 & 57.92 \\
    % bz:19200         & 1024 & 63.90 & 47.71 & 61.67 & \textbf{68.59} & 60.12 & 47.73 & 59.00 \\
    % w.o. Instruct    & 1024 & 70.55 & 53.00 & 76.77 & 68.58 & 64.91 & \textbf{50.01} & 63.40 \\
    BGE-\textit{pretrain}         & 1024 & 63.90 & 47.71 & 61.67 & \textbf{68.59} & 60.12 & 47.73 & 59.00 \\
    BGE w.o. pre-train   & 1024 & 62.56 & 48.06 & 61.66 & 67.89 & 61.25 & 46.82 & 58.62 \\
    BGE w.o. Instruct    & 1024 & 70.55 & 53.00 & 76.77 & 68.58 & 64.91 & \textbf{50.01} & 63.40 \\
    \hline
    BGE-\textit{finetune}  & 1024 & \textbf{71.53} & \textbf{54.98} & \textbf{78.94} & 68.32 & \textbf{65.11} & 48.39 & \textbf{63.96} \\
    \hline
    % \ChangeRT{1pt}
    \end{tabular}
    % \vspace{-5pt}
    %\caption{Analysis of the impacts introduced by C-MTP and training recipe.} 
    % \caption{\textbf{Ablation of the training data, \data{}, and the training recipe.}}
    % \vspace{-10pt}
    \label{tab:4}
\end{table*}

\subsection{General Evaluation}
We extensively evaluate \model{} against popular Chinese text embeddings on \benchmark{} as shown in \autoref{tab:3}.\footnote{Our \model{} models are named \textbf{BGE} in the tables.}, where we can make the following observations.

% our models in C-TEM is empirically competitive, consistently better, directly applicable
% advantage is most notable for retrieval, then comes re-rank, sts, pair-clf, the most common usages of embedding
% openai embedding is not chinese oriented by OK

First, our models outperform existing Chinese text embeddings by large margins. There is not only an overwhelming advantage in terms of the average performance, but also notable improvements for the majority of tasks in \benchmark{}. The biggest improvements are on the retrieval task followed by STS, pair classification, and re-ranking. Such aspects are the most common functionalities of text embeddings, which are intensively utilized in applications like search engines, open-domain question answering, and the retrieval augmentation of large language models. Although the advantages for classification and clustering tasks are not as obvious, our performances are still on par or slightly better than the other most competitive models. The above observations verify the strong generality of \model{}. \textit{Our models can be directly utilized to support different types of application scenarios.}

% larger scale -> better performance
% small is still competitive, provide user with flexibility 

Second, we observe performance growth resulting from the scaling up model size and embedding dimension. Particularly, the average performance improves from 58.28 to 63.96, when the embedding model is expanded from small to large. Besides the growth in average performance, there are also improvements across all the evaluation tasks. Compared to the other two baselines (Text2Vec, M3E), the impact of scaling up is more consistent and significant for our models. It is worth noting that our small model is still empirically competitive despite its highly reduced model size, where the average performance is even higher than the large-scale option of many existing models. As a result, \textit{it provides people with the flexibility to trade-off embedding quality and running efficiency}: people may resort to our large-scale embedding model to deal with high-precision usages, or switch to the small-scale one for high-throughput scenarios. 

Using the same recipe as the Chinese models, we also train a set of English text embedding models presented in \autoref{tab:english}. Besides, our English data was created and released together with C-MTP. It was the first time that such comprehensive training data was made publicly available. At the time of public release, our English BGE models achieved the state-of-the-art performance on the English MTEB benchmark~\citep{muennighoff2022mteb} across its 56 datasets. Although there were many strong competitors in the English community, such as E5~\cite{wang2022text}, SGPT~\cite{muennighoff2022sgpt}, GTE~\cite{li2023towards}, GTR~\cite{ni2021large}, and OpenAI Ada-002~\cite{neelakantan2022text}, we were able to notably advance the prior SOTA by an absolute 1.1 points in total average, which further verify the effectiveness of our data curation and training method.

\subsection{Detailed Analysis}

We investigate the detailed impact of \data{} and our \textbf{training recipe}. The corresponding experiment results are presented in \autoref{tab:4} and \autoref{tab:4.1}, respectively. 

First of all, we analyze the impact of our training data, \data{}. As mentioned, \data{} consists of two parts. 1) \data{} \textbf{(unlabeled)}, which is used for general-purpose fine-tuning; the model produced from this stage is called the intermediate checkpoint, denoted as \textbf{BGE-}\textit{pretrain}. 2) \data{} \textbf{(labeled)}, where the task-specific fine-tuning is further conducted on top of BGE-\textit{pretrain}; the model produced from this stage is called the final checkpoint, noted as \textbf{BGE}-\textit{finetune}. Based on our observations from the experimental result, both \data{} \textbf{(unlabeled)} and \data{} \textbf{(labeled)} substantially contribute to the embedding's quality. 

Regarding \data{} \textbf{(unlabeled)}, despite mostly being curated from unlabeled corpora, this dataset alone brings forth strong empirical performance for the embedding models trained on it. Compared with other baselines like Text2Vec, M3E, and OpenAI text embedding, BGE-\textit{pretrain} already achieves a higher average performance. A further look into the performances reveals more details. On one hand, \data{} \textbf{(unlabeled)} makes a major impact on the embedding's retrieval quality, where BGE-\textit{pretrain} notably outperforms the baselines in this attribute. On the other hand, the general capability of embedding is primarily established with \data{} \textbf{(unlabeled)}, as BGE-\textit{pretrain}'s performance is close to the baselines on the rest of the aspects, like STS and Clustering. \textit{This puts our embedding models in a very favorable position for further improvements.}

As for \data{} \textbf{(labeled)}, the dataset is much smaller but of better quality. With another round of fine-tuning on \data{} \textbf{(labeled)}, the empirical advantage is significantly expanded for the final checkpoint BGE-\textit{finetune}, where it gives rise to a jump in average performance from 59.0 (BGE-\textit{pretrain}) to 63.96 (BGE-\textit{finetune}). Knowing that the text pairs in \data{} \textbf{(labeled)} are mainly gathered from retrieval and NLI tasks, the most notable improvements are achieved on closely related tasks, namely retrieval, re-ranking, STS, and pair classification. On other tasks, it preserves or marginally improves performance. \textit{This indicates that a mixture of high-quality and diversified labeled data is able to bring forth substantial and comprehensive improvements for a pre-trained embedding model.}

% shows that BGE-\textit{pretrain}'s performance is close to the baselines on most of the aspects, like STS, Clustering, etc., meaning that the general capability can be primarily established with C-MTP (unlabeled). Moreover, BGE-\textit{pretrain} notably outperforms the baselines on the retrieval task, which reflects the value of MTP (unlabeled) in 

\begin{table}[t]
    \centering
    % \small
    % \footnotesize
    % \begin{tabular}{|p{1.0cm}|C{2.5cm}|C{2.5cm}|}
    \caption{\textbf{Impact of batch size.}}
    \begin{tabular}{|p{3.0cm}|C{1.2cm}|C{1.2cm}|C{1.2cm}|}
    % \ChangeRT{1pt} 
    \hline
    \backslashbox{Task}{Batch Size} & 256 & 2,048 & 19,200   \\
    \hline
    Retrieval & 57.25 & 60.96 & \textbf{63.90} \\
    STS       & 46.16 & 46.60 & \textbf{47.71} \\
    Pair CLF  & \textbf{62.02} & 61.91 & 61.67 \\
    CLF       & 65.71 & 67.42 & \textbf{68.59} \\
    Re-rank   & 58.59 & 59.98 & \textbf{60.12} \\
    Cluster   & \textbf{49.52} & 49.04 & 47.73 \\
    \hline
    Average   & 56.43 & 57.92 & \textbf{59.00} \\
    \hline
    % \ChangeRT{1pt}
    \end{tabular}
    % \vspace{-5pt}
    % \caption{\textbf{Impact of batch size.}}
    % \vspace{-10pt}
    \label{tab:4.1}
\end{table} 

\begin{table*}[t]
    \centering
    % \footnotesize
    % \small
    \caption{\textbf{Performance of English Models on MTEB.}}
    \begin{tabular}{|p{2.4cm}|C{1.0cm}|C{1.2cm}|C{1.2cm}|C{1.2cm}|C{1.2cm}|C{1.2cm}|C{1.2cm}|C{1.5cm}|C{1.2cm}|}
    % {|l|c|c|c|c|c|c|c|c|c|c}
\hline
% \toprule
Model Name & Dim. & Average & Retrieval & Cluster & Pair CLF & Re-rank & STS & Summarize & CLF \\ 
\hline
% \midrule
GTE (large) & 1024 & 63.13 & 52.22 & \textbf{46.84} & 85.00 & 59.13 & 83.35 & 31.66 & 73.33 \\
GTE (base) & 768 & 62.39 & 51.14 & 46.2 & 84.57 & 58.61 & 82.3 & 31.17 & 73.01 \\
E5 (large) & 1024 & 62.25 & 50.56 & 44.49 & 86.03 & 56.61 & 82.05 & 30.19 & 75.24 \\
Instructor-XL & 768 & 61.79 & 49.26 & 44.74 & 86.62 & 57.29 & 83.06 & 32.32 & 61.79 \\
E5 (base) & 768 & 61.5 & 50.29 & 43.80 & 85.73 & 55.91 & 81.05 & 30.28 & 73.84 \\
GTE (small) & 384 & 61.36 & 49.46 & 44.89 & 83.54 & 57.7 & 82.07 & 30.42 & 72.31 \\
OpenAI Ada 002 & 1536 & 60.99 & 49.25 & 45.9 & 84.89 & 56.32 & 80.97 & 30.8 & 70.93 \\
E5 (small) & 384 & 59.93 & 49.04 & 39.92 & 84.67 & 54.32 & 80.39 & 31.16 & 72.94 \\
ST5 (XXL) & 768 & 59.51 & 42.24 & 43.72 & 85.06 & 56.42 & 82.63 & 30.08 & 73.42 \\
MPNet (base) & 768 & 57.78 & 43.81 & 43.69 & 83.04 & 59.36 & 80.28 & 27.49 & 65.07 \\
SGPT Bloom (7.1B) & 4096 & 57.59 & 48.22 & 38.93 & 81.9 & 55.65 & 77.74 & \textbf{33.60} & 66.19  
\\
\hline
% \midrule 
{BGE (small)} & 384 & 62.17 &51.68 & 43.82 &  84.92 & 58.36 & 81.59 & 30.12 & 74.14 \\
{BGE (base)} & 768 & 63.55 & 53.25 &   45.77 & 86.55 & 58.86 & 82.4 & 31.07 & 75.53 \\
{BGE (large)} & 1024 & \textbf{64.23} & \textbf{54.29} & 46.08 & \textbf{87.12} & \textbf{60.03} & \textbf{83.11} & 31.61 & \textbf{75.97} \\
\hline
% \bottomrule
    \end{tabular}
    % \caption{\textbf{Performance of English Models on MTEB.}}
    \label{tab:english}
\end{table*}

We make further exploration about our \textbf{training recipe}, particularly the impact from contrastive learning, task-specific fine-tuning, and pre-training. 

One notable feature of our training recipe is that we adopt a large batch size for contrastive learning. According to previous studies, the learning of the embedding model may benefit from the increasing of negative samples \cite{izacard2021unsupervised,qu2020rocketqa,muennighoff2022sgpt}. Given our dependency on in-batch negative samples, the batch size needs to be expanded as much as possible. In our implementation, we use a compound strategy of gradient checkpointing and cross-device embedding sharing \cite{gao2021scaling}, which results in a maximum batch size of 19,200. By making a parallel comparison between bz: 256, 2028, 19,200, we \textit{observe consistent improvement in embedding quality with the expansion of batch size} (noted as bz). The most notable improvement is achieved in retrieval performance. This is likely due to the fact that retrieval is usually performed over a large database, where embeddings need to be highly discriminative. 

Another feature is the utilization of instructions during task-specific fine-tuning. The task-specific instruction serves as a hard prompt. It differentiates the embedding model's activation, which lets the model better accommodate a variety of different tasks. We perform the ablation study by removing this operation, noted as ``\textit{w.o. Instruct}''. Compared with this variation, the original method BGE-f gives rise to better average performance. Besides, there are more significant empirical advantages on retrieval, STS, pair classification, and re-rank. All these perspectives are closely related to the training data at the final stage, i.e. \data{} \textbf{(labeled)}, where the model is fine-tuned on a small group of tasks. This indicates that \textit{using instructions may substantially contribute to the quality of task-specific fine-tuning.}

One more characteristic is that we use a specifically pre-trained text encoder to train \model{}, rather than using common choices, like BERT~\cite{devlin2018bert} and RoBERTa~\cite{liu2019roberta}. To explore its impact, we replace the pre-trained text encoder with the widely used Chinese-RoBERTa\footnote{huggingface.co/hfl/chinese-roberta-wwm-ext-large}, noted as ``BGE w.o. pre-train''. According to the comparison between BGE-\textit{pretrain} and BGE w.o. pre-train, \textit{the using of pre-trained text encoder notably improves the retrieval capability, while preserving similar performances on other perspectives.}

\bibliographystyle{ACM-Reference-Format}
\bibliography{main}

\end{document}